\documentclass[letterpaper]{article} 
\usepackage[]{aaai25}  
\usepackage{times}  
\usepackage{helvet}  
\usepackage{courier}  
\usepackage[hyphens]{url}  
\usepackage{graphicx} 
\usepackage{natbib}  
\usepackage{amsmath}
\usepackage{amssymb}

\usepackage{caption} 
\usepackage{multirow}
\usepackage{algorithm}
\usepackage{algorithmic}
\usepackage{booktabs} %
\usepackage{array}
\usepackage{newfloat}
\usepackage{listings}
\DeclareCaptionStyle{ruled}{labelfont=normalfont,labelsep=colon,strut=off} 
\floatstyle{ruled}
\newfloat{listing}{tb}{lst}{}
\floatname{listing}{Listing}
\usepackage{bibentry}

\title{HM3: Hierarchical Multi-Objective Model Merging for Pretrained Models}
\author {
	Yu Zhou\textsuperscript{\rm 1},
	Xingyu Wu\textsuperscript{\rm 2},
	Jibin Wu\textsuperscript{\rm 2},
	Liang Feng\textsuperscript{\rm 3},
	Kay Chen Tan\textsuperscript{\rm 2},
}
\affiliations {
	\textsuperscript{\rm 1}Department of Computing, The Hong Kong Polytechnic University, Hong Kong SAR, China\\
	\textsuperscript{\rm 2}Department of Data Science and Artificial Intelligence, The Hong Kong Polytechnic University, Hong Kong SAR, China\\
	\textsuperscript{\rm 3}College of Computer Science, Chongqing University, Chongqing, China\\
	zy-yu.zhou@connect.ployu.hk, 
	\{xingy.wu, jibin.wu, kctan\}@polyu.edu.hk, liangf@cqu.edu.cn
}

\begin{document}

\maketitle

\begin{abstract}
Model merging is a technique that combines multiple large pretrained  models into a single model with enhanced performance and broader task adaptability. It has gained popularity in large pretrained model development due to its ability to bypass the need for original training data and further training processes. However, most existing model merging approaches focus solely on exploring the parameter space, merging models with identical architectures. Merging within the architecture space, despite its potential, remains in its early stages due to the vast search space and the challenges of layer compatibility. This paper marks a significant advance toward more flexible and comprehensive model merging techniques by modeling the architecture-space merging process as a reinforcement learning task. We train policy and value networks using offline sampling of weight vectors, which are then employed for the online optimization of merging strategies. Moreover, a multi-objective optimization paradigm is introduced to accommodate users' diverse task preferences, learning the Pareto front of optimal models to offer customized merging suggestions. Experimental results across multiple tasks, including text translation, mathematical reasoning, and code generation, validate the effectiveness and superiority of the proposed framework in model merging. The code will be made publicly available after the review process.
\end{abstract}

\section{Introduction}

\begin{figure*}[htb] 
	
	\center{\includegraphics[width=1\linewidth]  {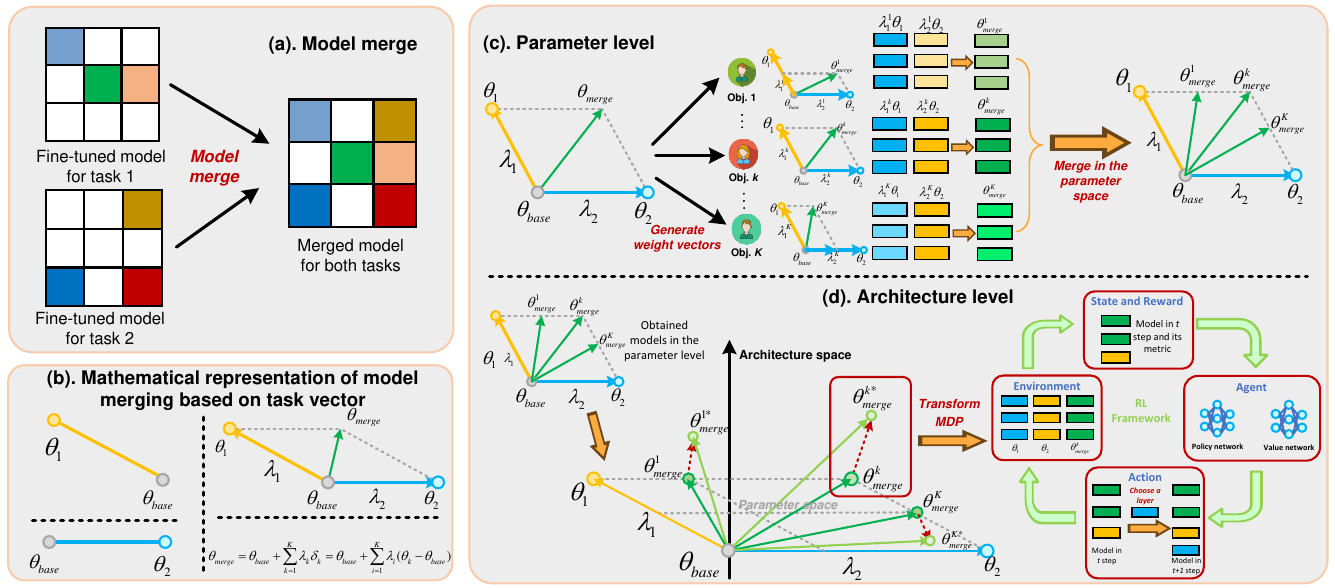}}\caption{Illustration of model merge and our proposed method. (a) illustrates the concept and process of model merging. (b) presents the mathematical representation of model merging named task vector. Based on the task vector, model merging is performed through task arithmetic. (c) and (d) demonstrate the model merging process of TM3 at the parameter and architecture levels, respectively. In (c), HM3 uniformly generates parameter vectors for different base models, and then optimizes the optimal parameters using a combination of the DARE method and Ties Merging. In (d), based on the model obtained in (c), HM3 employs a reinforcement learning strategy to reconstruct the model within a layer-granularity search space. } 
	\label{fig:framework} 
\end{figure*}

Recent advancements in large pretrained models have demonstrated remarkable performance and strong generalization abilities across various domains, such as natural language processing \cite{chang2024survey, zhou2024causalbench,wu2024evolutionary} and computer vision \cite{wang2024visionllm,radford2021learning}. Open-source communities have provided many pretrained models for various data types, as well as fine-tuned versions tailored to specific tasks. However, large fine-tuning  models is often a complex process that requires vast amounts of high-quality data and computational resources. To address the challenge of building foundational models capable of handling diverse tasks under limited computational resources, model merging has gained increasing attention \cite{jang2024model}. Model merging leverages existing pretrained models to flexibly transfer and integrate knowledge without requiring the original training data or additional model training \cite{white2016sampling}. This approach enables the creation of new models with higher generalization capabilities, suited to multiple tasks and scenarios. In recent years, model merging has become a simple yet popular method in pretrained models development, as illustrated in Fig. \ref{fig:framework}(a), with merged models showing significant potential on the Open LLM leaderboard \cite{myrzakhan2024open}. Current model merging methods primarily focus on merging models with the same architecture in the parameter space. In recent years, research in the parameter space has become quite extensive, including approaches like weight averaging \cite{matena2022merging}, Model Soup \cite{wortsman2022model}, Ties merging \cite{yadav2024ties}, and DARE (Drop And REscale) method \cite{yu2024language}. 

However, focusing solely on merging models within the parameter space significantly limits their practical utility. Models with different architectures exhibit broader diversity in representation capabilities and task types \cite{mellor2021neural,zoph2018learning}, potentially expanding the performance boundaries of merged models beyond those of a single architecture. Nevertheless, merging models across different architectures presents several practical challenges \cite{dong2021nats}, leading to limited research in this area. Firstly, architecture-level merging alters the computational logic of the model, necessitating the design of coordination strategies to ensure internal compatibility and seamless information flow within the new architecture. Moreover, jointly exploring both the parameter space and architecture space increases the problem's complexity \cite{9430615}, requiring well-defined search spaces and efficient search strategies to identify the optimal model configuration. Although evolutionary algorithms have been employed to search for optimal architectures \cite{akiba2024evolutionary}, they fail to reveal the mapping between architecture sequences and performance, making them unsuitable for handling the complex, high-dimensional problem of merging multiple models. Additionally, evolutionary processes are often one-time fusions, requiring a complete restart when faced with new problems, leading to significant computational consumption.

To merge models across both parameter and architecture levels and achieve reusable model merging schemes, this paper proposes a hierarchical model merging framework. In the parameter space, it uniformly assigns parameter vectors to different base models, with methods like DARE and Ties Merging \cite{huang2024emr,lu2024twin} optimizing the optimal parameters. Based on the optimal merged model obtained in the parameter space, a reinforcement learning (RL) strategy \cite{mnih2015human} reconstructs the model within a layer-granularity search space. The training process first samples rewards under different architectural states and uses them as supervision to train policy and value networks. A transformation matrix is constructed to ensure compatibility between layers, with its parameters jointly determined by the parameter distributions of the connected layers. The policy and value networks trained in the framework can be reused to predict optimal merging architectures and parameters for different tasks.

Compared to most existing methods, the proposed framework has another advantage of fully considering conflicts or interference across tasks by employing a multi-objective optimization manner \cite{tan2005multiobjective,10056413}. In practical applications, users may have different preferences and expectations for the merged model. For instance, a legal translation agency might prefer a model that excels in translating legal documents while maintaining adequate performance in legal reasoning and terminology consistency \cite{briva2024large}. Thus, different base models/tasks should be weighted distinctively. Unlike existing methods that treat all base models equally, the proposed hierarchical multi-objective model merging (HM3) assigns weight hyperparameters to different base models, essentially seeking the Pareto front of optimal merging solutions across multiple tasks rather than a single averaged model. The final merged result allows users to select the most suitable model based on their specific needs and trade-offs from the Pareto front. The main contributions of this paper are summarized as follows:

\begin{itemize}
\item By introducing reinforcement learning to guide the search, this paper presents the first reusable model merging method that explores the optimal model configurations in both the parameter and architecture spaces.
\item The paper adopts a multi-objective learning paradigm that allows users to prioritize the importance of multiple tasks based on task needs by searching for the Pareto front of merging strategy, enabling them to select the most suitable merged model.
\item The proposed method demonstrates state-of-the-art performance across various benchmarks, including text translation, mathematical reasoning, and code generation tasks, outperforming traditional model merging methods in effectiveness and efficiency.
\end{itemize}

\section{Background}\label{sec:pre}

Model merge refers to combining the parameters and features of multiple large pretrained  models to generate a unified model that can perform better in multiple tasks. Through model merge, the advantages of different models can be utilized to enhance the model's generalization and multi-task processing capabilities. In the general setting of model merge, given a set of $K$ tasks and the corresponding pretrained or fine-tuned models, whose parameters are denoted as $\{\pmb{\theta}_1, \pmb{\theta}_2, \ldots, \pmb{\theta}_K\}$. The goal of model merging is to combine these $K$ models into a single model that can effectively handle all $K$ tasks. It is important to note that these models are fine-tuned from the same base model with parameters $\pmb{\theta}_{base}$. The merging process can be represented as \cite{cong2024have}:
\begin{equation}
	\pmb{\theta}_{merge} = g_{merge}(\pmb{\theta}_{base}, \pmb{\theta}_1, \pmb{\theta}_2, \ldots, \pmb{\theta}_K)
\end{equation}
where $\pmb{\theta}_{merge}$ is the parameters of the merged model that can efficiently perform all $K$ tasks; and $g_{merge}$ represents the model merging method. The illustration of the mathematical representation of model merging is shown in Fig. \ref{fig:framework}(b). 

Among existing model merge methods, average merging \cite{matena2022merging} is a common approach, which constructs merged models by averaging parameters expressed as $\pmb{\theta}_{merge} =  \sum_{k=1}^{K} \frac{\pmb{\theta}_k}{K}$. Model soups \cite{wortsman2022model} generates a multifunctional composite merged pretrained model by simply linearly combining the parameters of multiple fine-tuned models, denoted as $\pmb{\theta}_{merge} = \sum_{k=1}^{K} \lambda_k \pmb{\theta}_k$. Task arithmetic \cite{ilharcoediting} uses predefined scaling coefficients to differentiate the importance of various models, which is described as $\pmb{\theta}_{merge} = \pmb{\theta}_{base} + \sum_{k=1}^{K} \lambda_k \delta_k = \pmb{\theta}_{base} + \sum_{k=1}^{K} \lambda_k (\pmb{\theta}_k-\pmb{\theta}_{base})$. Ties merging method \cite{yadav2024ties} addresses the task conflict problem in task arithmetic by pruning low-magnitude parameters, resolving sign discrepancies, and non-overlappingly merging parameters with consistent signs. DARE (Drop And REscale) merge method \cite{yu2024language} sets most delta parameters denoted as $\delta_k = \pmb{\theta}_k-\pmb{\theta}_{base}$ to zero and rescales the remaining embeddings to approximate the original embeddings. DARE, as a general plug-and-play technique, sparsifies delta parameters of multiple fine-tuned models of the same model architecture,  which reduces parameter interference and merges them into a single model. 
Recently, many other model merging methods have also emerged \cite{davari2023model,deep2024della,jang2024model}. For detailed information on the related work, please refer to \textbf{Appendix-A}.

\section{Method}

\subsection{Overview}
Existing approaches predominantly focus on optimization in the parameter space, whereas our approach simultaneously considers both the parameter and architecture space and aims to achieve a more effective merged model. Consequently, this paper seeks to optimize the parameters and architectures of pretrained models with the objective of obtaining a set of Pareto-optimal merged models, which not only exhibit superior performance but also cater to diverse user preferences. In general, we first provide a standard mathematical formulation for multi-objective model merge in parameter space $\mathcal{P}$ and architecture space $\mathcal{A}$:
\begin{subequations}
	\begin{align}
		\mathop { \min_{\pmb{\theta}, \alpha}}   f(\pmb{\theta}, \alpha)=&(f_1(\pmb{\theta}, \alpha), f_2(\pmb{\theta}, \alpha), \ldots, f_K(\pmb{\theta}, \alpha))\\
		\text{s.t.}  \quad \mathcal{C}1: &\pmb{\theta} \in \mathcal{P}, \label{contraint2}\\
		\quad \mathcal{C}2: &\alpha \in \mathcal{A}.
	\end{align}
\end{subequations}
where $f_k(\cdot)$ ($k\in\{1,2,\ldots,K\}$) represents the performance metric on the merged model for the $k$-th task, $\pmb{\theta}$ and $\alpha$ corresponds to the parameters and architecture of the merged model, respectively. 
The optimization objective, whether to maximize or minimize $f_k(\cdot)$, depends on the property of the metric. Additionally, $\mathcal{C}1$ ensures that the model parameters reside within the parameter space, and $\mathcal{C}2$ requires that the model architecture adheres to the architecture space.

In general, solving this problem requires advanced search algorithms to identify both model parameters and architecture. However, given the vast scale of pretrained models, the combined search space of parameters, and architecture is exceedingly large, presenting a significant challenge even for the most advanced search algorithms. To address this, we first transform the problem into a hierarchical optimization problem, including the parameter and architecture 
level, and design a hierarchical multi-objective model merge approach named HM3 to solve it.  At the parameter level, we focus on optimizing the parameters of large pretrained  models for different preferences and find a Pareto-approximate optimal model in the parameter space, while at the architecture level, we refine the architecture of pretrained models to enhance their performance based on the optimal merged models obtained at the parameter level. 


\subsection{Parameter Level}
In the parameter level, since we extend the concept of model merging within the parameter space to the multi-objective field, HM3 first generates weight vectors, and then HM3 is used to obtain the optimal merged model within the parameter space for each weigh vector. The illustration of this process is provided in Fig. \ref{fig:framework}(c), and the details are as follows.
\subsubsection{Weight vector generation}

For the fine-tuned models corresponding to multiple tasks, the process of model merging assigns to these fine-tuned models a weight vector denoted as $\pmb{\lambda} = (\lambda_1, \lambda_2, \ldots, \lambda_K)$, where $\lambda_k$ ($k\in\{1,2,\ldots,K\}$) represents the weight value based on the $k$-th model (this is also called the scaling factor in related works), and satisfies the  conditions:
\begin{equation}
	\sum_{k=1}^K \lambda_k = 1, \quad \lambda_k \geq 0 \quad \text{for } k = 1, 2, \ldots, K.
\end{equation}

In multi-objective model merge, we achieve this by uniformly generating $N$ weight vectors denoted as $\{\pmb{\lambda}^1, \pmb{\lambda}^2, \ldots, \pmb{\lambda}^N\}$ on the unit simplex method which is conceptually similar to the ideas of MOEA/D-based methods \cite{4358754,10098867} in multi-objective optimization. Each weight vector effectively represents a specific-preference model merging problem, and the set of $N$ weight vectors collectively defines the multi-objective model merging problem. Here, $N = \binom{K + q - 1}{K - 1}$,
where $K$ represents the number of objectives, and $q$ is used to control both the number and the distribution density of the generated weight vectors. After weight vector generation, we need to search the optimal merge model in the parameter space for each weight vector. 

\subsubsection{Optimization in the parameter space}In the parameter space, HM3 introduces the drop and rescale mechanism in \cite{yu2024language} to enhance the robustness of the merged model and then utilizes Ties merging to obtain the merged model in the parameter space. Specifically, we first operate by dropping weights, where certain elements within the weight vector are randomly set to zero, which corresponds to dropping specific components. For the $k$-th task,
we randomly retain the delta parameter with a probability $p$ (setting the others to zero) to create $\delta_k$, and then rescale the remaining parameters and obtain the new delta parameter $\delta_{k}^{DR} = \frac{\delta_k}{1 - p}$. After that, HM3 proceeds to Ties merging \cite{yadav2024ties} for each weight vector, which is designed to further refine the merged model by addressing issues such as redundant parameters and sign inconsistencies among different base models. For each weight vector, we first trim redundant parameters from $\delta_k^{DR}$  and then create an aggregated sign vector to resolve sign inconsistencies across different models. Finally, the disjoint components $\delta_k'$ from different tasks are merged to form the final merged model parameters.
\begin{equation}
	\pmb{\theta}_{para} = \pmb{\theta}_{base} + \pmb{\lambda} \sum_{k=1}^{K}\delta_k'
\end{equation}
For all weigh vectors $\{\pmb{\lambda}^1, \pmb{\lambda}^2, \ldots, \pmb{\lambda}^N\}$, we can obtain the corresponding optimal merged model denoted as $\{\pmb{\theta}_{para}^1, \pmb{\theta}_{para}^2, \ldots, \pmb{\theta}_{para}^N\}$ in the parameter space.

\subsection{Architecture Level}
Previous research has demonstrated the effectiveness and potential of employing search algorithms to optimize the model architecture for enhancing the performance of merged models \cite{akiba2024evolutionary}. The core idea is to explore the data flow space by reordering and combining layers from multiple models and then using a evolutionary search algorithm to find the optimal sequence of layers. However, as the number and size of models and layers increase, the search space becomes more complex which leads to a marked decrease in the search algorithm's efficiency. Due to its population-based nature, each round of evaluation in evolutionary search can consume a lot of computing resources. In addition, evolutionary search requires training from scratch for each weight vector. Therefore, the development of novel optimization strategies has become an urgent need.

Based on prior efforts \cite{akiba2024evolutionary}, we find that searching for the inference path can significantly reduce the search space compared to directly optimizing a model architecture. Therefore, HM3 in the architecture space first creates a token to traverse through the layers of multiple fine-tuned or merged models and finds the optimal $\alpha$ by searching for the optimal inference path. This inference path is a sequence of transformations of the layers and models followed by the token, which determines how the information is processed and ultimately affects the performance of the merged model. To this end, we aim to optimize the inference path and obtain the corresponding merged model of the optimal inference path for each weight vector, which is illustrated in Fig. \ref{fig:framework}(d). Specifically, over a set of $K+1$ models (i.e., $K$ fine-tuned models and one obtained optimal merged model in the parameter space parametered as $\pmb{\theta}_{para}$) for each weight vector, the optimal architecture $\alpha$ is considered as the optimal inference path represented as a sequence $I^t_{m,l}|_{t=1}^T$, where ${m,l}$ represents the $l$-th layer in the $m$-th model and $T$ is the total length of the inference path. 

In order to search the optimal inference path, HM3 needs to dynamically choose between layers in multiple models, which is essentially a multi-step decision problem, and the decision at each time step affects the choice of subsequent times steps and the final merged model's performance. This sequential decision-making process is inherently suited to being modeled as a Markov decision process (MDP), which allows RL methods to find an optimal solution \cite{10436092}. Therefore, we first convert the problem of searching the optimal $\alpha$ into an MDP and set up the key components of the MDP, which is the mathematical representation of the RL framework. Subsequently, HM3 constructs a novel RL optimization algorithm to solve the MDP and obtain the final merged model.



\subsubsection{Mathematical representation of RL framework}

During the model merging process, the token needs to dynamically choose between layers in multiple models and ultimately determine an inference path, which is essentially a multi-step decision problem. To this end, we first transform the model merge problem in the architecture space as an MDP, where each time step is the token from one layer in a model to another layer in the current or another model, and the merged model in the $t$-th step has $t$ layers. Additionally, this MDP consists of the following key components:

The state $s_t$ in the $t$-th step is defined as the current position of the token in the inference path. In the architecture level, we have $K$ fine-tuned model and one optimal merged model in the parameter space, each with $L$ layers. Therefore, the state can be expressed as:
\begin{equation}
	S_t = (m_t, l_t), m \in [1,K+1], l \in [1, (K+1)L]
\end{equation}
where $m_t$ denotes the index of the selected model and $l_t$ represents the index of the layer in the selected $m$-th model in the $t$-th step. 

The action $A_t$ in the $t$-th step is defined as the decision to transition to the next layer in the next model based on the current state, which can be represented as:
\begin{equation}
	A_t = (m_{t+1}, l_{t+1}), m \in [1,K+1], l \in [1, (K+1)L]
\end{equation}

The reward function quantifies the impact of the current action on the performance of the merged model. We define a combined reward in the $t$-th step as 	$R_t = R_t^{metric} + R_t^{path\_penalty}$, consisting of two components. The first component is the metric-based reward, reflecting the performance metric of the merged model, which is defined as
\begin{equation}
	R_t^{metric} = f(S_t, A_t)
\end{equation}
The second part is the path complexity penalty, which is to encourage the selection of shorter and more efficient inference paths. In this paper, path complexity is the length of the inference path, which is given as:
\begin{equation}
	R_t^{path\_penalty} = -\beta_1 t
\end{equation}
where $\beta_1$ is a constant to control the path length impact.

%
%

\subsubsection{RL algorithm}
In order to solve the MDP, HM3 utilizes a popular reinforcement learning method named a proximal policy optimization (PPO) \cite{schulman2017proximal,huang2024ppo} to optimize the inference path, which is a popular RL algorithm designed for high-dimensional and continuous action spaces, and PPO has an actor-critic framework including a policy network parametered as $\mu$ and a value network parametered as $\phi$.

The policy network with policy function $\pi_\mu(A_t | S_t)$ takes the current state $S_t$ as input and outputs the probability distribution over possible actions. The objective of the policy network is to maximize the expected reward:
\begin{equation}
	\max_\mu \mathbb{E}_{\pi_\mu} \left[ \sum_{t=0}^{T} \gamma^t R_t \right]
\end{equation}

The value network with value function $V_\phi(S_t)$ estimates the value of the current state, which is the expected cumulative reward:
\begin{equation}
	V_\phi(S_t) = \mathbb{E}_{\pi_\mu} \left[ G_t \mid S_t \right]
\end{equation}
where  $G_t$ is the cumulative reward starting from the $t$-th step  calculated as $G_t = \sum_{j=0}^{\infty} \gamma^j R_{t+j}$.

HM3 stabilizes policy optimization by constraining the update steps. The loss functions consist of policy loss, value loss, and overall loss. Specifically, the policy loss is given by:
\begin{equation}\label{eq:policy_loss}
	L^{CLIP}(\mu) = \mathbb{E}_t \left[ \min\left(\rho_t(\mu) \hat{A}_t, \text{clip}(\rho_t(\mu), 1 - \epsilon, 1 + \epsilon) \hat{A}_t\right)\right]
\end{equation}
where $\rho_t(\mu) = \frac{\pi_\mu(A_t | S_t)}{\pi_{\mu_{old}}(A_t | S_t)}$ is the ratio of the new and old policies, and $\hat{A}_t$ is the advantage function. In this paper, $\hat{A}_t$ in the $t$-th step is calculated by using the generalized advantage estimation: $\hat{A}_t = \sum_{i=0}^{\infty} (\gamma \beta_{A})^i \zeta_{t+i}$
where $\zeta_t = R_t + \gamma V(S_{t+1}; \phi_{iter}) - V(S_t; \phi_{iter})$ in the $iter$-th episode.

The value loss is defined as:
\begin{equation}\label{eq:value_loss}
	L^{VF}(\phi) = \mathbb{E}_t \left[ \left( V_{\phi}(S_t) - \hat{G}_t \right)^2 \right]
\end{equation}
where $\hat{G}_t$ is the target cumulative reward calculated by $\hat{G}_t=V_{\phi}(S_t)-\hat{A}_t$.

The overall loss function is a weighted combination of the policy and value losses, along with an entropy term to encourage exploration:
\begin{equation}
	L(\mu, \phi) = L^{CLIP}(\mu) + c_1 L^{VF}(\phi) - c_2 H(\pi_\mu)
\end{equation}
where $H(\pi_\mu)$ is the entropy of the policy, and $c_1$ and $c_2$ are coefficients.

To adapt to the varying input distributions between different layers, we introduce a feedforward multi-layer perceptron (MLP) network that generates a scaling matrix $W_{m,l}$. Specifically, the input to the MLP network consists of the current layer index pair $(m, l)$ and the time step $t$. The output is the corresponding $W_{m,l}$. The MLP network can be expressed as:
\begin{equation}
	W_{m,l} = \text{MLP}_{\mu}(m, l, t)
\end{equation}
where $\text{MLP}_{\mu}$ denotes the  MLP network and its parameters $\mu$ are optimized using PPO.

The overall algorithm is summarized as Algorithm \ref{alg:ppo_rl}. It begins by taking in a collection of weight vectors and the associated best-merged models from the parameter space (Line 1). Each weight vector introduces the fine-tuned models and the corresponding best-merged model (Line 3). The initial parameters for the policy network, value network, and MLP network are set up (Line 4). During each loop, the policy network interacts with the environment, creating a trajectory composed of state, action, and reward information (Line 6). The state at the current step and the chosen action are determined next (Lines 7-8). Subsequently, the reward for the current action is calculated based on the state within the merged model (Line 9). These rewards are then utilized to compute the advantage and target values (Line 10). The algorithm adjusts the policy network by enhancing the policy loss and refines the value network by minimizing the value loss (Lines 11-12). The process also involves calculating the scaling matrix and fine-tuning the MLP network through PPO (Line 13). Finally, it outputs the final set of optimized policy parameters and the sequence of actions that represent the optimal inference path tied to the value network (Line 16).

\begin{algorithm}[t]
	\caption{HM3 in the architecture space}\label{alg:ppo_rl}
	\begin{algorithmic}[1]
		\STATE \textbf{Input:} A set of weight vectors $\{\pmb{\lambda}^1, \pmb{\lambda}^2, \ldots, \pmb{\lambda}^N\}$ and their corresponding optimal merged models at the parameter space.
		
		\FOR{each weight vector $\pmb{\lambda}^n$ in $\{\pmb{\lambda}^1, \pmb{\lambda}^2, \ldots, \pmb{\lambda}^N\}$}
		\STATE Input $K$ fine-tuned models and the optimal merged model in the parameter space corresponding to the weight vector $\lambda_i$.
		\STATE Initialize the parameters of policy network as $\mu_0$, of value network as $\phi_0$, of MLP network as $\text{MLP}_{\mu0}$,
		\FOR{each iteration $iter = 1, 2, \dots, Max\_iter$}
		\STATE Sample the current policy \( \pi_{\mu_{iter}}(A_t | S_t) \) by interacting with the environment to generate a trajectory of length \( T \) as \( \{S_t, A_t, R_t, S_{t+1}\}_{t=1}^{T} \)
		\STATE Obtain the state $S_t = (m_t, l_t)$
		\STATE Select the action $A_t = (m_{t+1}, l_{t+1})$.
		\STATE Calculate the reward $R_t$ based on $A_t = (m_{t+1}, l_{t+1})$ and the merged model in the $t$-th step.
		\STATE Compute the advantage function $\hat{A}_t$ and $\hat{G}_t$.
		\STATE Update the policy network by maximizing \eqref{eq:policy_loss}
		\STATE Update the value network by minimizing \eqref{eq:value_loss}
		\STATE Compute the scaling matrix $W_{m,l}$, and update the MLP network parameters $\text{MLP}_{\mu}$.
		\ENDFOR
		\ENDFOR
		\STATE \textbf{Output:}  Optimal policy network parametered $\mu^*$ and the optimal inference path (i.e., the optimal sequence of actions) corresponding to the value network.
	\end{algorithmic}
\end{algorithm}

\section{Experiment}

\subsection{Experiment Setup}
The primary objective of our study is to develop a set of high-performance, multi-task merged models. To achieve this goal, we applied our proposed HM3 to a collection of fine-tuned models, including Llama-7B-Chat \cite{touvron2023llama}, WizardMath-7B \cite{luo2023wizardmath}, and CodeLlama-7B \cite{roziere2023code}, which are all fine-tuned versions of Llama-2-7B \cite{touvron2023llama} for specific tasks. By leveraging these fine-tuned models, we can employ our proposed HM3 to merge models excelled across multiple tasks. To evaluate the performance of the merged models by HM3, we perform three tasks, including language translation, mathematical reasoning, and code generation. For achieving these evaluations quickly and efficiently, we employed two popular pretrained model evaluation packages: lm-evaluation-harness \cite{eval-harness} for text translation and mathematical reasoning tasks, and bigcode-evaluation-harness \cite{bigcode-evaluation-harness} for code generation tasks. To further demonstrate the effectiveness and superiority of our method compared to other model merging methods, we utilized the mergekit package to merge models by using several merging methods, including Task Arithmetic, Ties, and DARE-Ties.

As for the setting of optimization in the parameter space, the number of weight vectors $N$ is set to 15, while all other parameters for parameter space optimization are configured with the default settings like \cite{yu2024language,yadav2024ties}. In the architecture space, the maximum number of iterations $Max\_iter$ is set to 1000, and the training of the policy network and value network starts after 200 iterations. 

\subsection{Tasks and Metrics}
\subsubsection{Translation Tasks}
To evaluate the multilingual translation capabilities of pretrained models, we leveraged a set of translation tasks in the lm-evaluation-harness package, including WMT14, WMT16 \cite{sennrich2016edinburgh},  and IWSLT2017 \cite{cettolo2017overview}. These tasks evaluate the model's translation accuracy and fluency across diverse language pairs. For all translation tasks, we use the "chrf" metric, which measures translation quality based on character n-gram precision and recall.


\subsubsection{GSM8K}
GSM8K \cite{cobbe2021training} is a dataset meticulously designed for mathematical problem-solving tasks, comprising over 8,000 high-quality problems that span from basic arithmetic to complex algebra. The primary objective of this dataset is to evaluate the model's reasoning and computational abilities when tackling structured mathematical problems. For evaluating the GSM8K dataset, we employ the "flexible match" metric, which allows for minor variations in the final answer. 

\subsubsection{HumanEval}
HumanEval \cite{chen2021evaluating} is a benchmark dataset proposed by OpenAI, specifically designed to evaluate code generation capabilities. The dataset comprises 164 programming problems, where each problem requires the model to generate a Python function based on a natural language description. The evaluation metric of HumanEval is pass@100. The model is allowed to generate up to 100 code solutions for each problem. This metric assesses whether at least one of these generated solutions passes all test cases. 

All the links to the used models, datasets, and evaluation platforms are provided in \textbf{Appendix-B}.

%
%

\subsection{Performance}
\subsubsection{Performance of multi-tasks}
For the evaluation of the merged model's performance, we randomly sampled a weight vector and compared the results of HM3 with those of single models and merged models obtained by other merging methods on text translation, mathematical reasoning, and code generation tasks. The experimental results were summarized in Table \ref{tab:1}. Llama-2-7B-Chat showed a strong balance in text translation. WizardMath-7B performed exceptionally well on the mathematical reasoning task, with a score of 40.79, likely because this model was fine-tuned specifically for math problems. It also achieved good performance on the text translation task with a score of 34.97 but showed poor performance on the code generation task. CodeLlama-7B demonstrated strong performance on code generation with a score of 42.20, which significantly exceeded the performance of the other two fine-tuned models. However, it performed the worst on text translation and mathematical reasoning. As for different merge methods, the task arithmetic method achieved scores of 31.30, 37.83, and 21.36 on text translation, mathematical reasoning, and code generation, respectively, indicating consistent improvement across all tasks. The Ties method showed a slight decline in performance on mathematical reasoning but maintained strong performance on the other two tasks. The combination of DARE and ties method matched the task arithmetic method in mathematical reasoning but slightly outperformed it in text translation and code generation. Our proposed HM3 achieved performance metrics of 44.68, 45.62, and 43.62 on text translation, mathematical reasoning, and code generation tasks, respectively. Notably, HM3 showed significant improvements in all three tasks compared to single models and other merging methods. Overall, while the performance of each model varied across different tasks, an effective model merging strategy significantly enhanced overall performance. HM3 consistently achieved the best performance across all metrics.

\begin{table}[t]		
	\centering
	\caption{Performance comparison of different merging methods of pretrained models}
			\scalebox{0.75}[0.75]{
	\begin{tabular}{ccccc}
				\toprule
				\textbf{Merging Methods} & \textbf{Source Models} & \textbf{Translation} & \textbf{Math} & \textbf{Code} \\
				
				\midrule
				Fine-tuned Model 1 & Llama-2-7B-Chat & 40.23 & 15.39 & 19.51 \\
				Fine-tuned Model 2 & WizardMath-7B & 34.97 & 40.79 & 20.73 \\
				Fine-tuned Model 3 & CodeLlama-7B & 33.86 & 12.89 & {42.20} \\
				\midrule
				Task Arithmetic & Model 1 + 2 + 3 & 31.30 & 37.83 & 21.36 \\
				Ties & Model 1 + 2 + 3 & 34.15 & 29.73 & 26.35 \\
				DARE + Ties & Model 1 + 2 + 3 & 33.93 & 38.20 & 28.20 \\
				\midrule
				HM3 & Model 1 + 2 + 3 & \textbf{44.68} & \textbf{45.62} & \textbf{43.62} \\
				\bottomrule
			\end{tabular}
	}
	\label{tab:1}%
\end{table}

\subsubsection{Performance of multi-objective model merge}
\begin{figure}[t] 
	\center{\includegraphics[width=0.7\linewidth]  {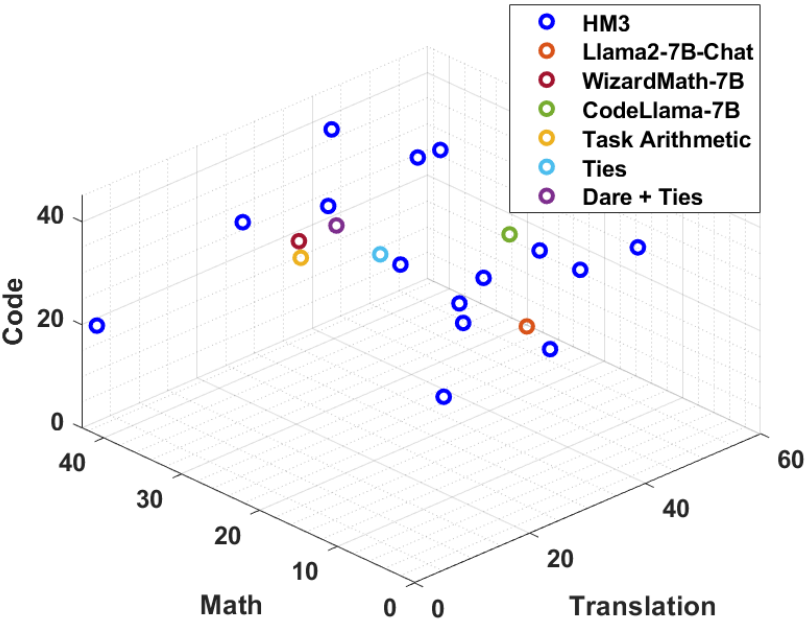}}\caption{Metrics in merged models for different model merging methods.} 
	\label{fig:1} 
\end{figure}

In this paper, we have initiated early explorations into multi-objective optimization for model merging to cater to different user preferences. By employing the HM3 method, we generated merged model sets corresponding to all weight vectors (all user preferences) and obtained the relevant metrics for each merged model within these sets. These metrics, along with those obtained from merged models generated by other methods, are depicted in Fig. \ref{fig:1}. As shown in Fig. \ref{fig:1}, our approach was capable of producing a set of Pareto-optimal merged models, along with their corresponding metrics, which provided valuable guidance for users to personalize their selection based on the specific needs of their tasks. In contrast, other methods were limited to generating only a single solution. Furthermore, we reported the Hypervolume (HV) value \cite{huband2003evolution, tan2005multiobjective}, and a higher HV value indicated a better solution set. The HV value of our proposed HM3 method was 1.6824.
\subsection{Performance of HM3 in different spaces}
\begin{figure}[t] 
	\center{\includegraphics[width=0.7\linewidth]  {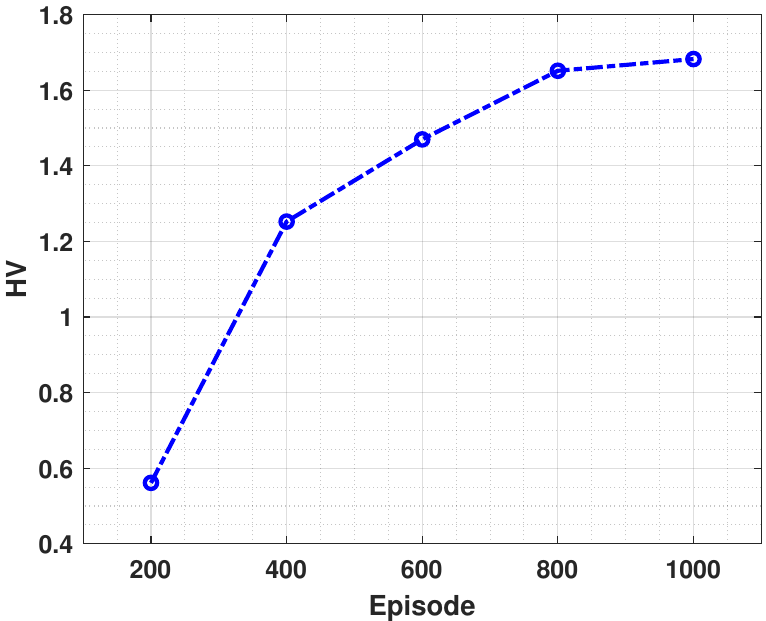}}\caption{HV in different episodes for HM3.} 
	\label{fig:2} 
\end{figure}

For the performance of HM3 in two spaces, we compared the following three methods: HM3, HM3 without architecture space optimization (i.e., HM3 w.o. arch), and HM3 without parameter space optimization short for HM3 w.o. para (i.e., the state in HM3 only included $K$ fine-tuned models). We conducted two experiments to demonstrate the impact of different levels on the performance of the merged models, and the results were summarized in Table \ref{tab:2}. The first experiment was for a single sampled weight vector. As shown in Table \ref{tab:2}, HM3 w.o. arch and HM3 w.o. para performed significantly worse than HM3 in mathematical reasoning and code generation tasks, while their performance in text translation tasks was closer to HM3. This may have been due to the base model already possessing text translation capabilities. HM3 w.o. arch performed better than HM3 w.o. para, which may have been because our method in the architecture space was still in the exploratory stage. To reduce the search space, we converted optimizing model architectures into optimizing inference paths, which might have reduced the performance of the merged models. The second experiment was for multiple weight vectors. We observed that HM3 still achieved the best HV, followed by HM3 w.o. arch, with the worst performance observed in HM3 w.o. para. These two experiments confirmed the effectiveness of HM3 in both parameter and architecture spaces and demonstrated that the performance of the merged model was related to both spaces. We also conducted an experiment on the HV values obtained by RL across different episodes, with the results provided in Fig. \ref{fig:2}. We observed that the performance of RL was very poor due to the initial policy network training. As training progressed beyond 200 episodes, HV values increased, stabilizing around 1000 episodes. We chose not to extend the episode count further due to computational cost considerations. Additionally, the remaining relevant experiments are provided in \textbf{Appendix-B}.
\begin{table}[t]
	\centering
	\caption{Performance of HM3 in different spaces}
				\scalebox{0.75}[0.75]{
	\begin{tabular}{c|ccc|c}
		\toprule
		\multirow{2}[3]{*}{Instance} & \multicolumn{3}{c}{Single objective} & Multi-objective \\
		\cmidrule{2-5}          & Translation & Math & Code & HV  \\
		\midrule
		HM3 w.o. para. opt. & 32.21 & 18.36 & 20.67 &1.2486 \\
		HM3 w.o. archi. opt. & 34.01 & 38.51 & 28.67 & 1.4564 \\
		HM3    & \textbf{44.68} & \textbf{45.62} & \textbf{43.62} & \textbf{1.6824} \\
		\bottomrule
	\end{tabular}%
}
	\label{tab:2}%
\end{table}%
\section{Conclusion}
In this paper, we have presented a novel hierarchical model merging framework (HM3) that addresses the limitations of existing model merging techniques by exploring both parameter and architecture spaces. HM3 introduces a reinforcement learning-based strategy to navigate the complex architecture-space merging process, thereby enabling the creation of more versatile and high-performing models. Through offline sampling of weight vectors, we trained policy and strategy networks that guide the online optimization of merging strategies. Furthermore, we incorporated a multi-objective optimization mechanism to address the varying preferences and requirements of users. By learning the Pareto front of optimal models, our framework provides customized merging solutions tailored to specific needs, thereby offering a more refined and user-centric approach to model merging. The effectiveness and superiority of our proposed framework have been validated through extensive experiments across multiple tasks, including text translation, mathematical reasoning, and code generation. The merged models produced by HM3 demonstrated significant improvements in performance compared to those generated by traditional merging methods. As model merging continues to evolve, future work can expand the HM3 to pretrained models with larger parameter scales, potentially leading to even more powerful and adaptable models.
\bigskip

\bibliography{aaai25}

\clearpage
\appendix
\onecolumn

{\centering\Huge{Appendix for ``HM3:~Hierarchical Multi-Objective Model Merging for Pretrained Models''}}



\section{Comprehensive related work about HM3}\label{app:causal_learning}
\subsection{Model Merge}
Model merge refers to combining the parameters and features of multiple pre-trained large models to generate a unified model that can perform better in multiple tasks. Through model merge, the advantages of different models can be utilized to enhance the model's generalization and multi-task processing capabilities. In the general setting of model merge, given a set of $K$ tasks and the corresponding pretrained or fine-tuned models, whose parameters are denoted as $\{\pmb{\theta}_1, \pmb{\theta}_2, \ldots, \pmb{\theta}_K\}$. The goal of model merging is to combine these $K$ models into a single model that can effectively handle all $K$ tasks. It is important to note that these models are fine-tuned from the same base model with parameters $\pmb{\theta}_{base}$. The merging process can be represented as \cite{cong2024have}:
\begin{equation}
	\pmb{\theta}_{merge} = g_{merge}(\pmb{\theta}_{base}, \pmb{\theta}_1, \pmb{\theta}_2, \ldots, \pmb{\theta}_K)
\end{equation}
where $\pmb{\theta}_{merge}$ is the parameters of the final merged model that can efficiently perform all $K$ tasks, and $g_{merge}$ represents the model merging method. The existing common and advanced model merge methods include model soup, task arithmetic, ties, and dare. Generally speaking, these methods focus on optimizing $\delta_k$ based on $\pmb{\theta}_{merge} = \pmb{\theta}_{base} + \sum_{k=1}^{K} \lambda_k \cdot \delta_k$ to obtain a better-merged model within the parameter space, where $\delta_k = \pmb{\theta}_k - \pmb{\theta}_{base}$ is the delta parameter or the task vector for the $k$-th task. Specifically,
\subsubsection{Model Soup}
Wortsman et al. \cite{wortsman2022model} proposed this method, which can simply linearly combine the parameters of $K$ fine-tuned models to generate multifunctional composite merge models, denoted as 
\begin{equation}
	\pmb{\theta}_{model\_soup} = \sum_{k=1}^{K} \lambda_k \cdot \pmb{\theta}_k
\end{equation}
where $\lambda_k$ is the scaling factor for generating a merged model. Generally speaking, for several model merging methods, $\lambda_k$ is also used to represent the scaling factor. In this paper, $\lambda_k$ not only indicates the scaling for the $k$-th fine-tuned LLM but also represents the weight (importance) of $k$-th task in the merging process. In this method, there is a special case named average merging, expressed as $\pmb{\theta}_{avg} = \frac{1}{K} \sum_{k=1}^{K} \pmb{\theta}_k$, which is also a popular model aggression method in federated learning \cite{mcmahan2017communication,10436092}.

\subsubsection{Task Arithmetic}
Ilharco et al. \cite{ilharcoediting} introduced a new paradigm for controlled model merging based on task vectors. This approach first constructs task vectors by subtracting parameters of the fine-tuned model for each task from the common base model, and then the performance differences between the fine-tuned model for $T_k$ and the base model can be reflected by the delta parameter as $\delta_k = \pmb{\theta}_k - \pmb{\theta}_{base}$, where $\delta_k$ is also known as the task vector for the $k$-th task. Therefore, this method uses the linear combination of multiple task vectors to generate the merged model, and the final merged model is calculated as follows:
\begin{equation}
	\pmb{\theta}_{task\_arithmetic} = \pmb{\theta}_{base} + \sum_{k=1}^{K} \lambda_k \cdot \delta_k
\end{equation}

The difference between model soup and task arithmetic merging method is that the latter requires knowledge of the base model (i.e., $\pmb{\theta}_{base}$).

\subsubsection{Ties Merging}
Since existing model merging methods often ignore the interference between parameters from different models, resulting in a significant deterioration in the performance of the merged model, Yadav et al. \cite{yadav2024ties} proposed a new model merging method, named ties, which aims to deal with interference caused by redundant parameter values and sign differences in specific parameter values between models. This method first creates a combined task vector $\delta_{{ties}}$, and the process of generating $\delta_{ties}$ includes trim, elect, and disjoint merge. 
Specifically,
This method first creates a combined task vector $\delta_{{ties}}$, and the process of generating $\delta_{ties}$ includes trim, elect, and disjoint merge. Specifically,
For each task $k$, the trim operation is to trim the redundant parameters from the delta parameter by keeping the top $n\%$ values and setting the remaining $(100 - n)\%$ parameters to zero. Then, the trimmed delta parameter for the $k$-th task can be obtained. Elect operation first creates an aggregated sign vector to resolve sign inconsistencies across different LLMs. For each parameter $p \in \{1, 2, \ldots, P\}$, separate the delta parameters by sign (+1 or -1) and calculate the total magnitude in the positive and negative directions. Then, the aggregated sign vector is assigned to the sign with the larger total magnitude.
For each $p$, the disjoint merge operation retains only the values from models where the sign matches the aggregated sign vector and calculates their average. Then, the final delta parameter $\delta_{ties}$ are calculated.
Finally, $\pmb{\theta}_{ties}$ is calculated as:
\begin{equation}
	\pmb{\theta}_{ties} = \pmb{\theta}_{base} + \lambda \delta_{ties}
\end{equation}

\subsubsection{DARE}
Yu et al. \cite{yu2024language} proposed a state-of-the-art pre-processing method aimed at improving the performance of merged models by increasing the sparsity of fine-tuned models. This method randomly drops the parameters of delta parameters with a probability $p$, and then rescales the remaining parameters by $1/(1 - p)$, then obtains updated delta parameters denoted as $\delta_{k}^{dare}$ for the $k$-th task.
\begin{equation}
	\delta_{k}^{dare}= \frac{\delta'_k}{1 - p}
\end{equation}
This method can combined with any model merging method. 
Incorporating DARE and the task arithmetic method, the model merging process is reformulated as: \begin{equation}
	\pmb{\theta}_{dare\_task\_arithmetic} = \pmb{\theta}_{base} + \sum_{k=1}^{K} \lambda_k \cdot \delta_k^{dare}
\end{equation}

\subsection{Multi-objective Optimization}
\subsubsection{Definition}
Generally, a multi-objective optimization problem can be formulated as:

\begin{equation}
	\min f(\pmb{x}) = (f_1(\pmb{x}), f_2(\pmb{x}), \ldots, f_K(\pmb{x})) \\
	\quad {s.t.} \quad \pmb{x} \in X,
\end{equation}
where $\pmb{x} = (x_1, x_2, \ldots, x_d)$ is a decision vector, and $f(\cdot): X \to Y$ represents $k$ objective functions. Here, $X$ denotes the decision space, and $Y$ denotes the objective space. To compare the quality of solutions obtained by the multi-objective problem, the concept of Pareto dominance is introduced.

Given two solutions $\pmb{x}_1$ and $\pmb{x}_2$ belonging to $X$, $\pmb{x}_1$ is said to Pareto dominate $\pmb{x}_2$ (denoted as $\pmb{x}_1 \prec \pmb{x}_2$) if and only if the following two conditions are satisfied:

\begin{enumerate}
	\item For all objectives $i \in \{1, 2, \ldots, K\}$, \(f_i(\pmb{x}_1) \leq f_i(\pmb{x}_2)\), meaning that \(\pmb{x}_1\) is not worse than \(\pmb{x}_2\) in every objective.
	\item There exists at least one objective $j \in \{1, 2, \ldots, m\}$ such that $f_j(\pmb{x}_1) < f_j(\pmb{x}_2)$, indicating that $\pmb{x}_1$ is strictly better than $\pmb{x}_2$ in at least one objective.
\end{enumerate}
A solution $\pmb{x}^* \in X$ is considered Pareto optimal if no other solution $\pmb{x} \in X$ Pareto dominates $\pmb{x}^*$. The set of all Pareto optimal solutions is known as the Pareto set:
\begin{equation}
	PS = \{\pmb{x} \in X \mid \nexists~\pmb{x}' \in X, \pmb{x}' \prec \pmb{x}\}
\end{equation}

The collection of objective vectors corresponding to the Pareto set is referred to as the Pareto front. The aim of multi-objective optimization is to approximate the Pareto set by identifying solutions that achieve both strong convergence and a diverse spread within the objective space.

In multi-objective optimization methods, since the true Pareto optimal solution set is unknown, we employ the commonly used metric called hypervolume (HV) \cite{tan2005multiobjective} to comprehensively assess the diversity and convergence of the generated approximate Pareto optimal solution set. Let a point set $P \subset \mathbb{R}^d$ and a reference point $\mathbf{r} \in \mathbb{R}^d$, where $d = 3$ is the number of optimization objectives. The HV of the set $P$ is computed as follows:
\begin{equation}
	\text{HV}(P, \mathbf{r}) = \mathcal{L}_e\left(\bigcup_{\mathbf{p} \in P} \left\{ \mathbf{q} \mid \mathbf{p} \preceq \mathbf{q} \preceq \mathbf{r} \right\}\right)
\end{equation}
where $\mathcal{L}_e(\cdot)$ represents the Lebesgue measure of a set: $\mathcal{L}_e(\mathcal{S}) = \int_{\mathbf{s} \in \mathcal{S}} \mathbf{1}_{\mathcal{S}}(\mathbf{s}) \, d\mathbf{s}$
Here, $\mathbf{1}_{\mathcal{S}}$ is the characteristic function of the objective space $\mathcal{S}$. If $\mathbf{s} \in \mathcal{S}$, then $\mathbf{1}_{\mathcal{S}}(\mathbf{s}) = 1$; otherwise, $ \mathbf{1}_{\mathcal{S}}(\mathbf{s}) = 0$. In the calculation of HV, the non-dominated solutions obtained by each algorithm are normalized using the same reference set, and the reference point is typically set at $(1, 1)$. It is important to note that a larger HV indicates a better approximation of the Pareto optimal solution set and, consequently, better performance of the corresponding multi-objective optimization method.

As for multi-objective optimization in model merging, there are two early explorations. The first paper \cite{li2024map} introduced a novel method called model merging with amortized Pareto fronts, which approximated evaluation metrics using a quadratic surrogate model derived from a set of pre-selected scaling coefficients. However, while this approach primarily focuses on reducing computational complexity, it does not thoroughly explore how to accurately obtain the Pareto-optimal merged model. The second paper \cite{li2024s} employed parallel multi-objective Bayesian optimization to systematically explore the parameter space for optimal merging configurations. However, these works are only in the early stages of exploration. They merely use multi-objective optimization methods to facilitate model merging but do not fully consider the multi-objective and multi-task characteristics inherent in the models during the merging process.

\subsection{Model Merge In the Data Flow Space}
Recent research \cite{akiba2024evolutionary} has explored the concept of knowledge distribution within language models, revealing promising avenues for model merging in the data flow space. Unlike traditional methods merging in the parameter space, model merging in the data flow space preserves the original layer weights while optimizing the paths of inference. This method allows tokens to transition across different layers of multiple models, such as moving from one layer in model A to other layers in model B, thereby enhancing the model’s versatility. Early investigations into merging in the data flow space primarily focused on serial connections with fixed, non-adaptive configurations. Specifically, given a set of models and a budget (representing the length of the inference path), the goal was to determine the optimal layer indices, which define the inference paths through the models. Due to the vast search space, researchers proposed a modified approach using an index array to manage the inclusion and exclusion of layers, effectively reducing the search space. The primary focus of this search process is to maintain the integrity of model parameters while optimizing the inference paths. Notably, model merge methods searching within the data flow space effectively aim to enhance the performance of the merged model by exploring different model architectures.

\section{Additional Results}\label{app:add_exp}
\subsection{Detail of Experimental Setup}
The primary objective of our study is to develop a set of high-performance, multi-task merged models. To achieve this goal, we applied our proposed HM3 to a collection of fine-tuned models, including Llama-7B-Chat \cite{touvron2023llama}, WizardMath-7B \cite{luo2023wizardmath}, and CodeLlama-7B \cite{roziere2023code}, which are all fine-tuned versions of Llama-2-7B \cite{touvron2023llama} for specific tasks. These fine-tuned models can be found in the following link:
\begin{itemize}
	\item Llama-7B-Chat: https://huggingface.co/meta-llama/Llama-2-7b-chat-hf;
	\item WizardMath-7B: https://huggingface.co/WizardLMTeam/WizardMath-7B;
	\item CodeLlama-7B: https://huggingface.co/codellama/CodeLlama-7b-hf;
	\item LLama-2-7B: https://huggingface.co/meta-llama/Llama-2-7b-hf.
\end{itemize}

By leveraging these fine-tuned LLMs, we can employ our proposed HM3 to merge LLMs excelled across multiple tasks. To evaluate the performance of the merged LLMs by HM3, we perform three tasks, including language translation, mathematical reasoning, and code generation. To achieve these evaluations quickly and efficiently, we employed two popular large model evaluation packages: lm-evaluation-harness \cite{eval-harness} for text translation and mathematical reasoning tasks and big code-evaluation-harness for code generation tasks. These evaluation packages can be found in the following link:
\begin{itemize}
	\item lm-evaluation-harness:~https://github.com/EleutherAI/lm-evaluation-harness;
	\item bigcode-evaluation-harness:~https://github.com/bigcode-project/bigcode-evaluation-harness.
\end{itemize}

To further demonstrate the effectiveness and superiority of our method compared to other model merging methods, we utilized the mergekit package \cite{goddard2024arcee}  to merge models by using several merging methods, including Task Arithmetic, TIES, and DARE-TIES. The mergekit package can be found at the following link:
\begin{itemize}
	\item mergekit:~https://github.com/arcee-ai/mergekit
\end{itemize}

Then, we introduce the specific datasets for text translation, math reasoning, and code generation tasks, as well as their metrics.
\subsubsection{Translation Tasks}
To evaluate the multilingual translation capabilities of LLMs, we leveraged a set of translation tasks in the lm-evaluation-harness package, including WMT14\footnote{https://www.statmt.org/wmt14/translation-task.html}, WMT16\footnote{https://www.statmt.org/wmt16/translation-task.html} \cite{sennrich2016edinburgh},  and IWSLT2017 \cite{cettolo2017overview}. These tasks evaluate the model's translation accuracy and fluency across diverse language pairs. For all translation tasks, we use the "chrf" metric, which measures translation quality based on character n-gram precision and recall.

\subsubsection{GSM8K}
GSM8K \cite{cobbe2021training} is a dataset meticulously designed for mathematical problem-solving tasks, comprising over 8,000 high-quality problems that span from basic arithmetic to complex algebra. The primary objective of this dataset is to evaluate the model's reasoning and computational abilities when tackling structured mathematical problems. For evaluating the GSM8K dataset, we employ the "flexible match" metric, which allows for minor variations in the final answer.  

\subsubsection{HumanEval}
HumanEval \cite{chen2021evaluating} is a benchmark dataset proposed by OpenAI, specifically designed to evaluate code generation capabilities. The dataset comprises 164 programming problems, where each problem requires the model to generate a Python function based on a natural language description. The evaluation metric of HumanEval is pass@100. The model is allowed to generate up to 100 code solutions for each problem. This metric assesses whether at least one of these generated solutions passes all test cases. 
\subsection{Effect of Different Number of Tasks}
\begin{figure}[htb] 
	\center{\includegraphics[width=0.4\linewidth]  {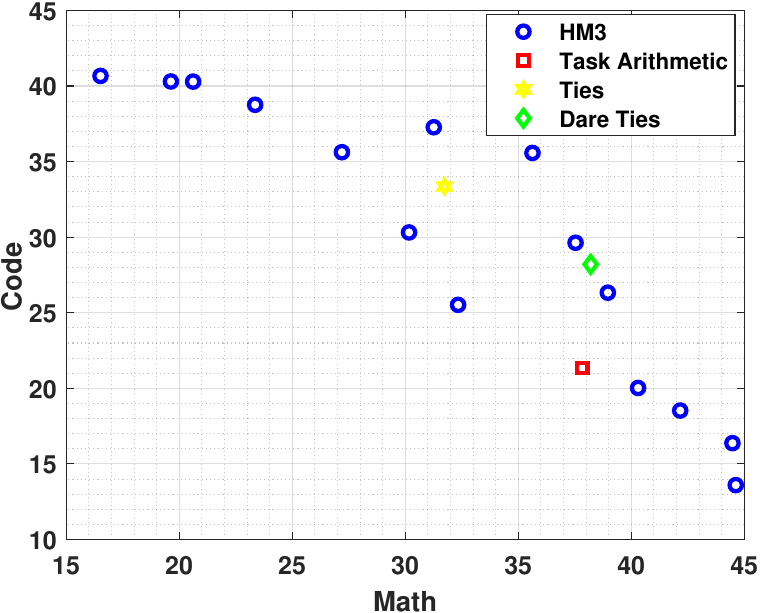}}\caption{The convergence of RL in the HM3 at the architecture space.} 
	\label{fig:Par_2} 
\end{figure}

In this subsection, we demonstrate the effectiveness of HM3 across different numbers of tasks. In the main text, we illustrated the effectiveness of HM3 on three tasks, and here we provide evidence of its effectiveness on two tasks, namely code generation and mathematical reasoning. The experimental results are presented in Fig. \ref{fig:Par_2}, which clearly illustrates the significant advantages of HM3. Specifically, HM3 is capable of generating a Pareto optimal set of solutions that excel not only in terms of parameter optimization but also in architectural configurations. The blue circles in the figure represent HM3, showing that its solutions are distributed across the entire performance curve, forming a comprehensive Pareto frontier that reflects optimal trade-offs under different conditions. In contrast, other methods, such as Task Arithmetic, Ties, and Dare Ties, are limited to generating a single optimal solution solely in the parameter space, as indicated by the red squares, yellow stars, and green diamonds, respectively. It is evident that the solution sets produced by these methods are confined to narrower regions of performance, lacking the diversity and flexibility that HM3 provides. Consequently, HM3 not only demonstrates the ability to explore the parameter space but also effectively leverages architectural optimizations to achieve a more comprehensive enhancement in performance.

\subsection{Effectiveness of Vision Tasks}
To further demonstrate the effectiveness of HM3, we constructed an image classification task based on the framework established by Ilharco \cite{ilharco2022patching,ilharcoediting}, and evaluated the performance of HM3 on various vision tasks. Specifically, we employed two variants of the Vision Transformer CLIP model \cite{radford2021learning}, namely ViT-B/32 and ViT-L/14, as visual encoders \cite{dosovitskiy2020image}. The visual encoders were then fine-tuned on five different tasks derived from the works of \cite{ilharco2022patching, ilharcoediting} and \cite{radford2021learning}, while keeping the text encoder unchanged. These tasks span a diverse range of classification domains, including remote sensing, traffic sign recognition, and satellite imagery. The datasets used in these experiments include DTD \cite{cimpoi2014describing},  GTSRB \cite{stallkamp2011german}, RESISC45 \cite{cheng2017remote}, SUN397 \cite{xiao2016sun}, and SVHN \cite{netzer2011reading}. The accuracy performance of different model merging methods on ViT-B/32 and ViT-L/14 across various datasets are detailed in Tables \ref{tab:1} and \ref{tab:2}, respectively. As shown in Table 1\ref{tab:1}, the HM3 method consistently outperforms other approaches on ViT-B/32, achieving an average accuracy of 66.91\%, which represents a significant improvement. Specifically, HM3 achieved 77.21\% and 77.62\% on the EuroSAT and SVHN datasets, respectively, and recorded a 68.21\% accuracy on the GTSRB dataset, surpassing other methods. Although its performance on the DTD dataset is slightly lower compared to the other tasks, it still outperforms the Ties and Task Arithmetic methods. Table \ref{tab:2} summarizes the performance of different merging methods on the ViT-L/14 model across various vision tasks. The table presents the accuracy results for four merging methods: Task Arithmetic, Ties, Dare Ties, and HM3, evaluated on several datasets, including SUN397, RESISC45, SVHN, GTSRB, and DTD. The results indicate that the HM3 method consistently achieved the best performance across most tasks, with particularly high accuracy on the SVHN and GTSRB datasets, reaching 90.48\% and 83.43\%, respectively. Notably, HM3 achieved an overall average accuracy of 80.30\% across all datasets, significantly outperforming the other merging methods. This demonstrates that the HM3 method excels at enhancing model performance when faced with diverse vision tasks, particularly in more complex classification challenges. These results in vision tasks further illustrate the superiority of HM3, particularly in effectively improving classification accuracy when handling challenging vision datasets.

\begin{table}[tbp]
	\centering
	\caption{Performance of merging ViT-B-32 model on vision tasks.}
	\begin{tabular}{lcccccc}
		\toprule
		Method & Average & SUN397 & RESISC45 & SVHN  & GTSRB & DTD \\
		\midrule
		Task Arithmetic & 64.67  & 61.41  & 72.42  & 73.60  & 66.12  & 49.82  \\
		Ties  & 63.75  & 62.34  & 71.49  & 73.68  & 62.69  & 48.52  \\
		Dare Ties & 64.99  & 60.22  & 71.60  & 76.56  & 65.94  & 50.60  \\
		HM3   & 66.91  & 63.22  & 73.27  & 77.62  & 68.21  & 52.22  \\
		\bottomrule
	\end{tabular}%
	\label{tab:1}%
\end{table}%

\begin{table}[tbp]
	\centering
	\caption{Performance of merging ViT-L-14 model on vision tasks.}
	\begin{tabular}{lcccccc}
		\toprule
		Method & Average & SUN397 & RESISC45 & SVHN  & GTSRB & DTD \\
		\midrule
		Task Arithmetic & 74.03  & 69.56  & 83.60  & 80.51  & 70.58  & 65.88  \\
		Ties  & 75.52  & 68.53  & 81.89  & 87.42  & 81.72  & 58.07  \\
		Dare Ties & 79.26  & 72.07  & 87.19  & 88.03  & 84.50  & 64.49  \\
		HM3   & 80.30  & 72.85  & 88.00  & 90.48  & 83.43  & 66.72  \\
		\bottomrule
	\end{tabular}%
	\label{tab:2}%
\end{table}%

\subsection{Effectiveness of RL}

\begin{figure}[htb] 
	\center{\includegraphics[width=0.4\linewidth]  {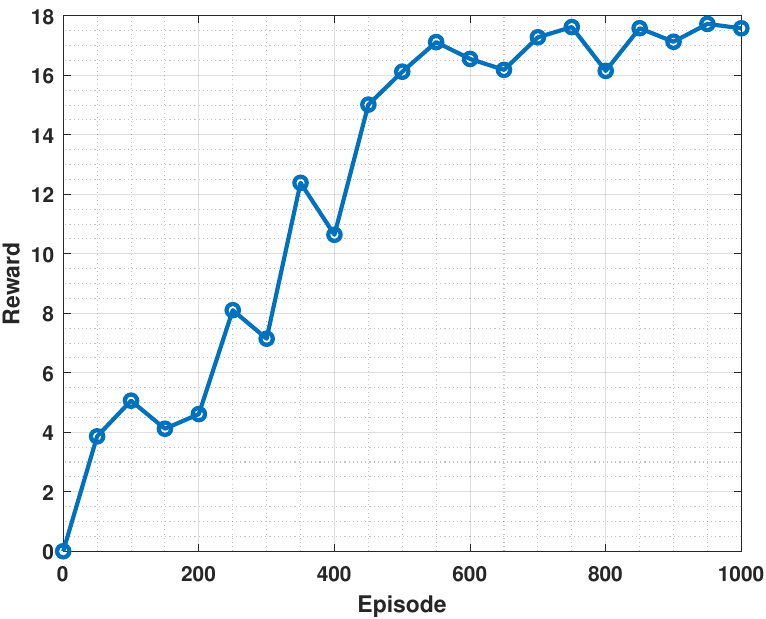}}\caption{The convergence of RL in the HM3 at the architecture space.} 
	\label{fig:RL_conv} 
\end{figure}

In this subsection, we delve into the convergence of TM3. Specifically, we randomly sample a weight vector and observe the obtained reward when merging models on text translation, mathematical reasoning, and code generation tasks. The experimental results are illustrated in Fig. \ref{fig:RL_conv}.  As shown in Fig. \ref{fig:RL_conv}, the overall reward increases progressively as the number of training episodes increases. During the first 200 episodes, the growth in reward was relatively slow, which is attributed to PPO's exploration phase, where HM3 had not yet accumulated sufficient experience and the policy network had not been trained. However, after episode 200, with the introduction of the experience replay mechanism, the reward begins to rise significantly, indicating that the algorithm is gradually learning from past experiences and improving its policy. As training continues, the reward shows a more stable upward trend and eventually converges to a value close to 18 around the 1000th episode. HM3 can effectively leverage past experiences to optimize its policy and achieve convergence.

\clearpage
\vspace{0.1in}
\bibliography{aaai25}
\end{document}